\documentclass{article}

\usepackage{arxiv}

\usepackage[utf8]{inputenc} %
\usepackage[T1]{fontenc}    %
\usepackage{hyperref}       %
\usepackage{url}            %
\usepackage{booktabs}       %
\usepackage{amsfonts}       %
\usepackage{nicefrac}       %
\usepackage{microtype}      %
\usepackage{lipsum}         %
\usepackage{todonotes}

\newcommand{\argmin}{\mathop{\rm argmin}}
\newcommand{\argmax}{\mathop{\rm argmax}}

\usepackage{url}            %
\usepackage{booktabs}       %
\usepackage{multirow}    
\usepackage{amsfonts}       %
\usepackage{nicefrac}       %
\usepackage{microtype}      %
\usepackage[numbers]{natbib}
\usepackage{enumerate}
\usepackage{hhline}
\usepackage{makecell}
\usepackage{pifont}

\usepackage{graphicx} %
\usepackage{caption}
\usepackage{subcaption}
\usepackage{amsmath}
\usepackage{amsthm}
\usepackage{amssymb}
\usepackage{tikz}
\usepackage{xcolor}
\usetikzlibrary{arrows}

\allowdisplaybreaks

\usepackage{mathrsfs}

\usepackage{algorithm}
\usepackage{algpseudocode}
\usepackage{hyperref}
\usepackage{bm}

\allowdisplaybreaks

\newcommand{\lnorm}{\left\Vert}
\newcommand{\rnorm}{\right\Vert}

\newcommand{\real}{\mathbb{R}}

\newcommand{\expect}{\mathbb{E}}

\usepackage{cleveref}
\crefname{thm}{Theorem}{Theorems}
\crefname{lem}{Lemma}{Lemmas}
\crefname{cor}{Corollary}{Corollaries}
\crefname{prop}{Proposition}{Propositions}
\crefname{asmp}{Assumption}{Assumptions}
\crefname{defn}{Definition}{Definitions}
\crefname{oracle}{Oracle}{Oracles}
\crefname{fact}{Fact}{Facts}
\crefname{conj}{Conjecture}{Conjectures}
\crefname{rem}{Remark}{Remarks}
\crefname{example}{Example}{Examples}
\crefname{condition}{Condition}{Conditions}
\crefname{exercise}{Exercise}{Exercises}
\crefname{algorithm}{Algorithm}{Algorithms}
\crefname{table}{Table}{Tables}
\crefname{figure}{Figure}{Figures}
\crefname{section}{Section}{Sections}
\crefname{subsection}{Section}{Sections}
\crefname{appendix}{Appendix}{Appendices}
\crefname{message}{Message}{Messages}

\definecolor{red}{rgb}{1, 0, 0}

\definecolor{green}{rgb}{0, 1, 0}

\definecolor{blue}{rgb}{0, 0, 1}

\definecolor{orange}{rgb}{1, 0.4, 0.0}

\input{math_commands}

\newcommand{\ucb}{\texttt{UCB}}
\newcommand{\piE}{\pi^{\operatorname{E}}}

\newcommand{\bonus}{\texttt{bonus}}
\newcommand{\critic}{\operatorname{critic}}
\newcommand{\actor}{\operatorname{actor}}

\newcommand{\meanstd}[2]{$#1 {\scriptscriptstyle \pm #2}$}

\hypersetup{
    colorlinks=true,
    citecolor=blue,
    linkcolor=blue,
}

\title{Deploying Offline Reinforcement Learning with Human Feedback}

\author{Ziniu Li\thanks{This work is done when Ziniu Li works as an intern in Tencent AI Lab.} \\
	The Chinese University of Hong Kong, Shenzhen\\
        \texttt{ziniuli@link.cuhk.edu.cn}
	\And Ke Xu, Liu Liu, Lanqing Li, Deheng Ye, Peilin Zhao\thanks{Corresponding author.}
	 \\
	Tencent AI Lab\\
	\texttt{ \{kaylakxu, leonliuliu, lanqingli, dericye, masonzhao\}@tencent.com} \\
}

\renewcommand{\headeright}{}
\renewcommand{\undertitle}{}
\renewcommand{\shorttitle}{Deploying Offline Reinforcement Learning with Human Feedback}

\begin{document}
\maketitle

\begin{abstract}
Reinforcement learning (RL) has shown promise for decision-making tasks in real-world applications. One practical framework involves training parameterized policy models from an offline dataset and subsequently deploying them in an online environment. However, this approach can be risky since the offline training may not be perfect, leading to poor performance of the RL models that may take dangerous actions. To address this issue, we propose an alternative framework that involves a human supervising the RL models and providing additional feedback in the online deployment phase.  We formalize this online deployment problem and develop two approaches. The first approach uses model selection and the upper confidence bound algorithm to adaptively select a model to deploy from a candidate set of trained offline RL models. The second approach involves fine-tuning the model in the online deployment phase when a supervision signal arrives. We demonstrate the effectiveness of these approaches for robot locomotion control and traffic light control tasks through empirical validation.
\end{abstract}

\setcounter{footnote}{0}

\section{Introduction}

Reinforcement learning (RL) offers a systematic approach to tackle sequential decision-making tasks \cite{sutton2018reinforcement}. RL methods utilize Markov Decision Processes (MDPs) to model the tasks \cite{puterman2014markov}, enabling agents to interact with an environment and enhance decision-making by maximizing long-term returns. Thanks to powerful neural networks, RL methods have achieved outstanding performance, surpassing even master-level expertise in various domains \cite{mnih2015human,DBLP:conf/aaai/YeLSSZWYYWGCYZS20, silver2017mastering, degrave2022magnetic, roveda2022q, chen2022inhomogeneous, DBLP:journals/tkde/ZhangZWLHT22, ouyang2022training}.

A popular RL framework for real-world applications involves two key steps. The first step is training parameterized policy models using an offline dataset that has previously been collected by specific behavior policies. The second step is deploying these trained policy models in an online environment. In recent years, significant efforts have been dedicated to training offline RL models \cite{fujimoto2019offpolicy, kumar2019stabilizing, kumar2020conservative, wu2020behavior, kostrikov2021offline, DBLP:conf/sigir/GaoXZLWYZ22}. The primary challenge in this approach is the lack of further data collection in the offline setting, which requires the agent to consider the epistemic uncertainty (i.e., subjective uncertainty due to limited samples) when optimizing policies. Consequently, various methods have been proposed and evaluated, which can be found in \cite{levine2020offline, fu2020d4rl} and their respective references. 

Despite its benefits, offline training may not be perfect due to various factors such as dataset quality and hyperparameter choices. As a result, trained models may suffer from overfitting, leading to poorer generalization performance in new scenarios. This is particularly concerning when deploying RL methods in the online phase, as models may take dangerous actions and unexpected results may occur. It is worth noting that the issue of overfitting is widely recognized in the machine learning community, and techniques such as cross-validation and early stopping have been proposed to evaluate model performance before deployment \citep{shalev2014understanding}. However, these methods often fail in RL due to the distributional shift problem \citep{ross2010efficient}. In other words, the training and validation distributions may differ significantly in decision-making tasks, making it challenging to assess the offline models' performance without conducting online experiments.

Apart from the evaluation issue mentioned earlier, another concern that arises in industrial applications (such as power system control, autonomous driving, and traffic light control) is the importance of safety and ethics \cite{garcia2015survey}. However, incorporating these factors into the training phase can be challenging as designing proper reward/penalty functions for them requires a significant amount of engineering effort \cite{pieter04apprentice}. Fortunately, in many applications, an expert policy (i.e., a human operator) can supervise the deployed RL model and provide feedback. For example, in the case of autonomous driving, people may have different preferences about the control system's decisions. Some may prioritize safety and comfort, while others may prioritize driving efficiency. In such scenarios, it is challenging to consider each person's preferences in the offline training phase, as feedback is only available in the online deployment phase. Therefore, offline RL methods may not perform well in this setting as they are not adaptive during online deployment. Although studies have emerged in the offline-to-online RL setting \cite{xu2021safely, schrittwieser2021online, lee2022offline}, these works have not considered online human feedback, which is crucial in practical applications.

\begin{figure}[htbp]
    \centering
    \includegraphics[width=0.55\linewidth]{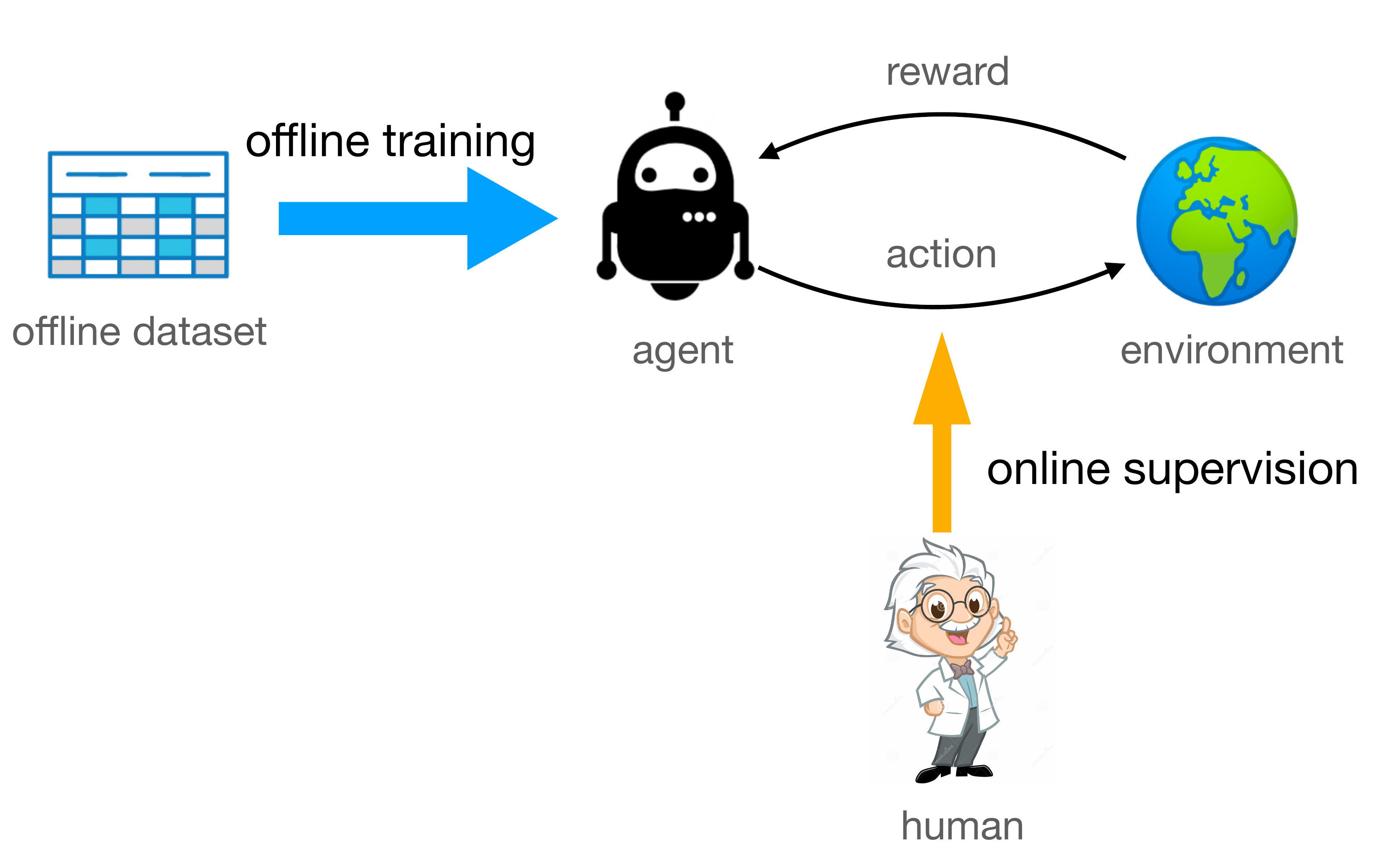}
    \caption{The framework of online deployment with human feedback. In our framework, a human expert supervises the RL models and provides additional feedback to improve their performance.}
    \label{fig:framework}
\end{figure}

After taking the above-mentioned considerations into account, it becomes essential to improve the performance of trained RL models during online deployment, particularly when human feedback is available. Please see \cref{fig:framework} for illustration. In this context, our objective is not only to maximize the environment return defined by each task but also to ensure that the decisions made by RL models are in line with what human experts expect. Thus, this paper proposes to maximize the concept of \emph{online score}, which integrates both the environment return and human feedback (further elaborated in \cref{subsec:deployment_of_rl_models}).

In this manuscript, we formalize the problem of online deployment with human feedback and propose two approaches to  maximize the online score. The first approach is based on model selection, where we assume there are $N$ pre-trained offline models, but their online scores are unknown in advance. To determine which model can achieve the highest online score with minimal trials, we propose using the upper confidence bound (UCB) algorithm \cite{lai1985asymptotically, auer02bandit}. The UCB algorithm estimates the online score of each offline model optimistically and adaptively selects the one to deploy, taking into account the stochastic and uncertain nature of the environment.

For the same online deployment problem, our second approach is based on fine-tuning. In this scenario, we assume that we only have access to one specific offline RL model, but we can improve its performance by fine-tuning it in the online deployment phase using human feedback. Unlike the first approach, the expert provides direct suggestions on action selection, allowing us to improve the model's performance. To leverage the human feedback, we develop imitation-learning-based methods \citep{hussein2017survey} that penalize the discrepancy between the model's and the human's decisions to improve the online score.

We conduct experiments on two tasks: robotics locomotion control \cite{duan2016benchmark} and traffic light control \cite{wei2018intellilight}. The goal of the first task is to train a robot to perform locomotion behaviors like humans, while in the second task, the objective is to control traffic lights to avoid congestion. We evaluate the performance of our proposed methods on these tasks. Specifically, we show that in the case of model selection, our approach can identify the best model from a candidate set with only about 100 trials. In the case of fine-tuning, we demonstrate that the online performance can be improved significantly with no more than 200 trials.

This paper is structured as follows. In Section \ref{sec:related_work}, we review prior research in the field. Section \ref{sec:problem_formulation} provides the necessary background and problem formulation. We then present our proposed methods for model selection and fine-tuning in Sections \ref{sec:online_model_selection} and \ref{sec:online_model_fine_tuning}, respectively. Finally, we present the numerical results in Section \ref{sec:experiments}.

\section{Related Work}
\label{sec:related_work}

In this section, we provide an overview of previous research related to the topic of this paper.

\textbf{Offline Reinforcement Learning.} Offline reinforcement learning algorithms aim to train an effective policy using a dataset that has been previously collected. Over the past few decades, offline RL (also known as batch RL) has been extensively studied in terms of algorithm design \cite{ernst2005tree, fujimoto2019offpolicy, kumar2019stabilizing, kumar2020conservative, wu2020behavior, kostrikov2021offline} and theoretical analysis \cite{szepesvari2005finite, munos2008finite, chen2019information, xie2021batch}. These works demonstrate that if the dataset has wide coverage and a small concentration coefficient in relation to the optimal policy, it is possible to accurately solve the Bellman equation using finite samples. In practical applications, the relevant theory studies suggest to consider the epistemic uncertainty in policy optimization. For example, BCQ \citep{fujimoto2019offpolicy} limits the action range to improve the policy, BRAC \citep{wu2020behavior} employs KL regularization during policy optimization, and CQL \citep{kumar2020conservative} penalizes Q-values for out-of-distribution actions.

It should be noted that prior studies on offline reinforcement learning algorithms have a significant limitation. These algorithms often employ online evaluation to optimize architectures or find suitable hyper-parameters, which is impractical due to the high cost of online evaluation. In real-world applications, hyperparameter-free heuristics such as selecting the action with the highest Q-value \citep{garg2020batch} are commonly used, despite the lack of theoretical guarantees. In our experiments, we will compare the performance of our proposed framework with such heuristics.

\textbf{Online Deployment.} Several recent studies have focused on the online deployment problem. For example, \citep{xu2021safely} proposed a meta-episodic algorithm that addresses the exploration uncertainty issue and ensures uniformly conservative exploration. \citep{schrittwieser2021online} employed model-based policy and value improvement operators to compute new training targets on existing data points. Additionally, \citep{lee2022offline} proposed a balanced experience replay scheme to address the online distribution shift issue. However, none of these works considered human feedback during the online deployment phase. In our framework, we consider two types of deployment plans: model selection and fine-tuning, and we review related works in the sequel.

\textbf{Model Selection.} Model selection is well studied in the supervised learning literature \cite{shalev2014understanding, mohri2018foundations}. Since we only have access to finite samples in practice, the generalization gap must be carefully considered when training an offline model.  In supervised learning, training and testing data are independently and identically drawn from the same distribution. Various techniques such as early stopping and $n$-fold cross-validation  have been developed to address overfitting in the training phase \citep{shalev2014understanding}. Nevertheless, reinforcement learning poses a unique challenge to model selection as training and testing data are not from the same distribution due to distributional shift \cite{levine2020offline}.

In contrast, model selection in online learning is well-studied in the literature, particularly in the context of multi-arm bandit (MAB) problems \cite{banditalgo}. In MAB, the learner seeks to identify the optimal arm based on partial and bandit feedback. In our scenario, we face a similar problem with partial and bandit feedback, making it suitable for MAB solutions. The upper confidence bound (UCB) algorithm is a popular approach for MAB, which provides nearly minimax optimal performance for regret minimization \cite{lai1985asymptotically, auer02bandit}. Therefore, we propose implementing the UCB algorithm for effective model selection in reinforcement learning.

\textbf{Fine-Tuning.} Fine-tuning is a widely used technique in deep learning, especially in the field of transfer learning \cite{goodfellow2016book}. It involves taking a pre-trained model and further refining it for downstream tasks \cite{yosinski2014how, kornblith2019do, chen2020simple}. In the realm of reinforcement learning (RL), researchers have explored the use of imitation learning approaches \cite{hussein2017survey} to initialize a model using human demonstrations, and then improve it further with online RL methods. One example of this is the DQfD algorithm proposed by \cite{hester2018demonstration}, which combines temporal difference updates with supervised classification of the demonstrator's actions. Another algorithm, LOKI, introduced by \cite{cheng2018fast}, starts with a few iterations of imitation learning before switching to a policy gradient RL method. A Bayesian formulation using prior information to fine-tune is considered in \citep{li2022hyperdqn}.

\section{Problem Formulation}
\label{sec:problem_formulation}

In this section, we first introduce the background of reinforcement learning (RL) in \cref{subsec:background}. Subsequently, we formalize the problem of online deployment with human feedback in \cref{subsec:deployment_of_rl_models}.

\subsection{Background}
\label{subsec:background}

\textbf{Markov Decision Processes.} A standard tool to study reinforcement learning is the Markov Decision Process (MDP) \cite{puterman2014markov}, which can be described by a tuple $(\gS, \gA, p, r, \rho, \gamma)$. Here $\gS$ and $\gA$ are the state and action space, respectively. Moreover, $p$ is the system transition function; i.e., $p(s^\prime|s, a)$ determines the probability of the next state $s^\prime$ condition on the current state-action pair $(s, a)$. $r: \gS \times \gA \rar \real$ specifies the reward signal and $\rho(\cdot)$ is the initial state distribution. Finally, $\gamma \in (0, 1)$ is the discounted factor in computing the long-term return. 

For a deterministic\footnote{Note that there always exists a deterministic policy that can achieve the optimal return \cite{puterman2014markov}, so it does not lose generality to consider the deterministic policies. } policy $\pi: \gS \rar \gA$, its expected long-term return is denoted by 
\begin{align*}
    V(\pi) &:= \expect\bigg[ \sum_{t=0}^{\infty} \gamma^{t} r(s_t, a_t) \mid s_0 \sim \rho, a_t = \pi(s_t),  s_{t+1} \sim p(\cdot|s_t, a_t), \forall t \geq 0 \bigg].
\end{align*}
To further measure the quality of policy $\pi$, $Q$-value function is introduced:
\begin{align*}
    Q^{\pi}(s, a) := &\expect\bigg[ \sum_{t=0}^{\infty} \gamma^{t} r(s_t, a_t) \mid  s_0 = s, a_0 = a; a_t = \pi(s_t), s_t \sim p(\cdot|s_t, a_t), \forall t \geq 1 \bigg],
\end{align*}
i.e., the expected long-term return starting from $(s, a)$. It is well-known that the optimal $Q$-value function $Q^{\star}$ satisfies the Bellman optimality equation:
\begin{align}   \label{eq:optimal_q}
    Q^{\star}(s, a) = r(s, a) + \gamma \expect_{s^\prime \sim p(\cdot|s, a)} \ls \max_{a^\prime} Q^{\star}(s^\prime, a^\prime) \rs.
\end{align}
The optimal policy $\pi^{\star}$ is defined as the greedy policy with respect to $Q^{\star}$, i.e., $\pi^{\star}(s) =  \argmax_{a \in \gA} Q^{\star}(s, a)$. In this manuscript, we mainly consider model-free approaches, so when we mention an RL model, we mean a $Q$-value function or the greedy policy associated with this $Q$-value function.

\textbf{Offline Reinforcement Learning.} In the framework of RL, the transition function is assumed to be unknown, so Equation \eqref{eq:optimal_q} cannot be directly solved. Instead, RL methods typically have access to samples obtained from environments in an online or offline manner. In this manuscript, we consider the offline scenario, where a dataset $\gD = \{(s, a, r, s^\prime)\}$ is provided to train RL models \cite{ernst2005tree}. In a simple form where the action space is finite, offline RL methods aim to minimize the Bellman error from finite samples:
\begin{equation}    \label{eq:td_learning}
\begin{split}
   \theta_{\critic} =  \argmin_{\theta} \sum_{(s, a, r, s^\prime) \in \gD} \bigg( &Q_{\theta}(s, a) - r(s, a)  - \gamma \max_{a^\prime} Q_{\theta}(s^\prime, a^\prime) \bigg)^2,
\end{split}
\end{equation}
where $\theta_{\critic}$ is the parameter of $Q$-value function $Q_{\theta_{\critic}}$ (in the context of actor-critic methods \cite{konda1999actor}, the $Q$-value function is also called a critic). Then, the policy can be extracted by greedily optimizing $Q_{\theta_{\critic}}$, i.e., $\pi(s) = \argmax_{a \in \gA} Q_{\theta_{\critic}}(s, a)$. 

For applications with continuous action control, the above training method cannot be applied as the maximization over $\gA$ cannot be directly implemented. In this case, we need an additional actor network $\pi_{\theta_{\actor}}$ to extract the greedy policy from $Q_{\theta_{\critic}}$:
\begin{align*}
   \theta_{\actor} = \argmax_{\theta} \sum_{s \in \gD} Q_{\theta_{\critic}}(s, \pi_{\theta} (s)).
\end{align*}
where $\theta_{\actor}$ is the parameter of the policy $\pi_{\theta_{\actor}}$ (in the context of actor-critic methods, the policy is also called an actor). Accordingly, the optimization problem \eqref{eq:td_learning} becomes
\begin{align*}
   \theta_{\critic} = \argmin_{\theta} \sum_{(s, a, r, s^\prime) \in \gD} \bigg( &Q_{\theta}(s, a) - r(s, a)  - \gamma Q_{\theta}(s^\prime, \pi_{\theta_{\actor}}(s^\prime)) \bigg)^2.
\end{align*}

Advanced offline RL methods additionally consider the epistemic uncertainty in the above optimization;  please refer to \cite{levine2020offline} and references therein. 

\subsection{Deployment of Offline RL Models}
\label{subsec:deployment_of_rl_models}

In this section, we explore the challenges involved in deploying trained RL models, which have significant implications for real-world applications. Although a well-trained offline RL model is expected to perform well in online deployment, this is often not the case. One of the primary reasons for this is the difficulty in accurately assessing the offline RL model's performance in the offline scenario. In supervised learning, we can use a validation dataset to choose a model that can generalize well. However, estimating the actual performance of an offline RL model using a validation dataset is challenging due to the distributional shift problem.

Another critical issue in deploying an offline RL model is safety \cite{garcia2015survey}. To address this concern, a human operator often supervises the trained RL model in the online phase to ensure that it does not take dangerous actions. Typically, this human operator has an expert policy $\piE$. If the RL agent deviates significantly from the expert policy, it incurs a penalty. Therefore, it is desirable for the trained RL model to balance the environment's return and the expert policy's expectation.

In this manuscript, we propose the expected online score as a metric to evaluate the performance of RL models in the online deployment phase, taking into account the challenges discussed earlier. The online score is defined as follows:
\begin{align} \label{eq:online_score}
S = \expect\bigg[ & \alpha_1 \cdot \sum_{t=1}^{T} r(s_t, \pi(s_t)) - \alpha_2 \cdot \sum_{t=1}^{T} \1 \lb \pi(s_t) \ne \piE(s_t) \rb \bigg]   \quad \text{(discrete action control)}. 
\end{align}
Here, $T$ is the maximum trajectory length of an episode, $\alpha_1 > 0$ and $\alpha_2 > 0$ are scaling factors, and $\1\lb \cdot \rb$ is the indicator function. The expectation is taken over the randomness in environment transitions. For continuous action control tasks, the second term in \eqref{eq:online_score} is too restrictive, so we introduce a relaxation that considers the squared distance between $\pi(s_t)$ and $\piE(s_t)$, with a tolerance parameter $\tau > 0$. That is, the second term in \eqref{eq:online_score} is replaced with 
\begin{align} \label{eq:online_score_1}
   S = \expect\bigg[ & \alpha_1 \cdot \sum_{t=1}^{T} r(s_t, \pi(s_t)) - \alpha_2 \cdot \sum_{t=1}^{T} \1 \lb \lnorm  \pi(s_t) - \piE(s_t) \rnorm^2 > \tau \rb \bigg]   \quad \text{(continuous action control)}.
\end{align}
We note that in both equations \eqref{eq:online_score} and \eqref{eq:online_score_1}, the second term acts as a constraint by quantifying the degree of disagreement between the RL models and the human expert.

\section{Online Model Selection}
\label{sec:online_model_selection}

In this section, we propose a model-selection-based approach to determine the best offline RL model for online deployment. Suppose we have trained $N$ different offline models using various methods such as different random seeds, training techniques, and hyperparameters. However, it is unclear which model would perform the best in online deployment. Therefore, we aim to select the optimal model by conducting multiple online deployment trials. This task can be viewed as an online model selection problem.

In this manuscript, we focus on the (cumulative) regret criterion as a model selection algorithm. The (cumulative) regret criterion is defined as follows:
\begin{align}
\label{eq:regret}
\text{regret}_K = \sum_{k=1}^{K} \left( S_{k} - S^{\star} \right),
\end{align}
where $S_k$ is the online score obtained when deploying a specific model in iteration $k$. $S^{\star}$ represents the expected best score that can be obtained by deploying the optimal offline model in hindsight, i.e., $S^{\star} = \max_{i = 1, \ldots, N} S^{i}$, where $S^{i}$ is the online score of the $i$-th model. Ideally, we want the model selection algorithm to identify the optimal model quickly, leading to sublinear growth of the regret.

It is worth noting that we only deploy one specific model at a time and receive its corresponding feedback. Thus, we are faced with a bandit feedback setting where the feedback information is partial, leading to an exploration-and-exploitation dilemma \cite{banditalgo}. This means that we must try each offline model multiple times (i.e., exploration) before identifying the optimal one for online deployment (i.e., exploitation). To balance the trade-off between exploration and exploitation adaptively, we can employ well-known UCB (upper confidence bound) strategies \cite{lai1985asymptotically, auer02bandit}, which construct an optimistic estimate of the feedback to guide exploration. Specifically, in each iteration $k$, we can use the following rule to make decisions:
\begin{align}   \label{eq:ucb_term}
     \argmax_{i} \ucb_k^{i} := \widehat{S^{i}_k} + \beta \cdot \bonus^{i}_k,
\end{align}
where $\widehat{S^{i}_k}$ is the estimation of the feedback for decision $i$ (explained later), $\beta > 0$ is the scaling factor, and $\bonus^{i}_k = \sqrt{1/n^{i}_k}$, where $n^{i}_k$ is the number of times decision $i$ has been attempted up to iteration $k$. Although this UCB strategy is a greedy approach, it can achieve sublinear regret due to the optimistic estimation \cite{banditalgo}. Intuitively, the bonus term encourages exploration, ensuring that each decision is tested sufficiently.

In our scenario, each decision refers to a specific offline model, and each iteration corresponds to a rollout of the policy, i.e., an episode. The feedback obtained in each iteration is a noisy score that combines the long-term environment return with a penalty for violating the human's intents, as follows:
\begin{align*}
    s_k = &\alpha_1 \cdot \sum_{t=1}^{T} r(s_t, a_t) - \alpha_2 \cdot \sum_{t=1}^{T} \1 \lb \pi(s_t) \ne \piE(s_t) \rb.
\end{align*}
Here, $\alpha_1$ and $\alpha_2$ are scaling factors defined previously, and $\piE$ is the expert policy. A similar formulation can be obtained for continuous action control tasks by replacing the hard constraint with the soft constraint. Note that $s_k$ is a random variable due to the randomness of environment transitions, and we have $S_k = \expect[s_k]$. 

The online score combines the environment rewards and human preferences for the agent's actions. Trained offline models can have dramatically different online scores. Reward-pursuit models, which have no particular constraints in offline policy optimization, may generate actions that human believe are risky and dangerous. Thus, such models receive a large penalty in the online deployment phase and a low online score. Conversely, conservative models with explicit constraints in offline policy optimization may not achieve excellent environment-defined performance, and radical experts may not like them, resulting in a low online score. In both cases, note that we do not know how the models will perform in the online deployment phase.

We have outlined the procedure for applying the UCB strategy to select offline models in our problem in \cref{algo:ucb}. In each iteration, we use the UCB strategy to select the most suitable offline model. The model index is denoted by $i_k$ in iteration $k$. We compute $\widehat{S^{i}_k}$ (appeared in \eqref{eq:ucb_term}) by the empirical mean: 
\begin{align}   \label{eq:hat_S}
    \widehat{S^{i}_k} = \frac{\sum_{k^\prime=1}^{k} s_{k^\prime} \1 (i_{k^\prime} = k) }{n_k^{i}},
\end{align}
where $n_k^{i}$ is the total times of deploying the $i$-th model. In \cref{algo:ucb}, we use $X_k^{j}$ to compute the numerator in \eqref{eq:hat_S}.

Although \cref{algo:ucb} is straightforward, it has been shown to achieve good practical performance and provides reasonable theoretical guarantees. According to the literature \cite{banditalgo}, the cumulative regret of \cref{algo:fine_tune} scales proportionally to $\widetilde{\gO}(\sqrt{NK})$, by properly choosing $\beta$. In practice, a constant value of $\beta$ usually performs well. Note that as $K$ goes to $\infty$, we have the averaged regret $\widetilde{\gO}(\sqrt{NK}/K) \rar 0$. The theory also implies that utilizing prior knowledge to select suitable candidate policies with small $N$ can significantly minimize regret.

\begin{algorithm}[htbp]
\begin{algorithmic}[1]
\caption{UCB for online model selection} \label{algo:ucb}
\Require{$N$: number of offline models, $\beta$: exploration coefficient, and offline models $M^1, \cdots, M^N$.}
\State{Initialize $X_0 \in \real^{N} \lar 0, n_0 \in \real^{N} \lar 0$.}
\For{iteration $k = 1, 2, \cdots$}
\If{$k \leq N$}
\State{$i_k = k$.}
\Else
\State{$i_k = \argmax_{i} \frac{X_k^{i}}{n_k^{i}}+ \beta \cdot \sqrt{\frac{1}{n_k^{i}}}$.} 
\EndIf
\State{Deploy the model $M^{i_k}$ and receive the score $s_{k}$.}
\For{each $j = 1, \cdots, N$}
\If{$j = i_{k}$}
\State{Update: $n_k^{j} \lar n_{k-1}^{j} +  1$ and $X_k^{j} \lar X^{j}_{k-1} +  s_{k}$.}
\Else
\State{Update: $n_k^{j} \lar n_{k-1}^{j}$ and $X_k^{j} \lar X^{j}_{k-1}$.}
\EndIf
\EndFor
\EndFor
\end{algorithmic}
\end{algorithm}

As a meta-algorithm, \cref{algo:ucb} can be applied to both discrete and continuous action control tasks. It is important to note that although UCB includes an exploration phase in the model selection process, it differs significantly from online exploration in a standard RL framework. In \cref{algo:ucb}, we consider only well-trained offline models and test them in the online phase, ensuring the quality of the exploration behavior. On the other hand, in a standard online RL framework, agents may attempt dangerous or harmful actions to explore, which can lead to unexpected results in some applications.

An advantage of \cref{algo:ucb} is its minimal computation complexity in the online phase, as it only requires storing a few vectors and updating them with simple calculations. However, the quality of the candidate set determines the online performance of this selection problem. If the quality of the $N$ offline RL models is poor, the final performance of \cref{algo:ucb} may not be acceptable. In such cases, fine-tuning can be used to further improve the performance of the offline models, and we will discuss this method in the following sections.

\section{Online Model Fine-Tuning}
\label{sec:online_model_fine_tuning}

In this section, we explore fine-tuning approaches to improve the quality of offline models during online deployment. Our inspiration for this method comes from the fine-tuning of deep neural networks in downstream tasks \cite{yosinski2014how, kornblith2019do, chen2020simple}, as well as fine-tuning of deep RL models trained from human demonstrations \cite{hester2018demonstration, cheng2018fast, aytar2018playing}. Specifically, we consider a scenario where a human expert policy can override the model's action when the expert's decision deviates significantly from the model's action. We log these events in the form of $(s, a, r, s')$, where $a$ is the expert's decision. Later, we extract samples from the log and construct a dataset $\gD^{o} = {(s, a, r, s')}$, which we use to fine-tune the models. Depending on whether the action space is continuous or discrete, we employ two different fine-tuning methods.

\textbf{Continuous action control.} In this scenario, we have two models: an actor model $\pi_{\theta_{\actor}}$ and a critic model $Q_{\theta_{\critic}}$ (refer to \cref{subsec:background}). It is natural to first optimize the critic by minimizing the Bellman error:
\begin{equation}  \label{eq:fine_tune_2}
\begin{split}
      \theta_{\critic}  = \argmin_{\theta} &\sum_{(s, a, r, s^\prime) \in \gD^{o}} \bigg( Q_{\theta}(s, a) - r(s, a)  - \gamma Q_{\theta}(s^\prime, \pi_{\theta_{\actor}}(s^\prime)) \bigg)^2.
\end{split}
\end{equation}
Then, we can improve the actor by maximizing its $Q$-value. To effectively follow the expert's guidance, we additionally train the actor by a mean-squared-error between the model's output and the expert's action (c.f. the second term in \eqref{eq:fine_tune_1}). This loss function is inspired by imitation learning theory \cite{xu2020error}. As a result, the trained actor may perform well in maximizing the environment return and following the expert, achieving a high score defined in \eqref{eq:online_score}.  
\begin{equation}
\begin{split}
\theta_{\actor} = \argmax_{\theta}  &\sum_{s \in \gD^{o}} Q_{\theta_{\critic}}(s, \pi_{\theta}(s))  -\sum_{(s, a) \in \gD^{o}} \lnorm \pi_{\theta}(s) - a \rnorm_2^2.   \label{eq:fine_tune_1}   
\end{split}
\end{equation}

\textbf{Discrete action control.}  Different from the previous case, there is no explicit component of policy or actor in the discrete action control applications. Instead, we only have a $Q$-value function. Hence, we cannot directly apply the above approach. Following \cite{hester2018demonstration}, we consider a margin-based loss function to increase the gap between the expert's action and the other actions. Concretely, the margin $\Delta > 0$ is a hyper-parameter, and this margin-based loss function incentivizes the expert action value $Q_{\theta_{\critic}}(s, a)$ is at least larger than $Q(s, a^\prime)$ for $a^\prime \ne a$ by a margin $\Delta$. In this way, the greedy policy is more likely to select the expert action $a$.
\begin{equation}    \label{eq:fine_tune_discrete}
\begin{split}
   \theta_{\critic} =  \argmin_{\theta} \sum_{(s, a, r, s^\prime) \in \gD^{o}} &\bigg\{ \bigg( Q_{\theta}(s, a) - r(s, a)   - \gamma Q_{\theta}(s^\prime, \pi_{\actor}(s^\prime)) \bigg)^2 \\
   &+ \max_{a^\prime} \left[ Q_{\theta}(s, a^\prime) + \ell_{\Delta}(a, a^\prime) - Q_{\theta}(s, a) \right] \bigg\}.
\end{split}
\end{equation}
where $\ell_{\Delta}(a, a^\prime) = \Delta$ if $a \ne a^\prime$ and $0$ otherwise.

In \cref{algo:fine_tune}, we present our proposed approach based on fine-tuning. However, it's important to note that in real-world scenarios, decisions must be made in real-time. Therefore, it's crucial to record only those state-action pairs where the expert is strongly dissatisfied with the decision made by the RL model. Otherwise, the log would become too large, leading to a significant computation load for the fine-tuning process. We address this issue in Lines 6-9 of \cref{algo:fine_tune}.

\begin{algorithm}[htbp]
\begin{algorithmic}[1]
\caption{Online Fine-Tuning} \label{algo:fine_tune}
\Require{Trained offline model.}
\State{Initialize $\gD^{o} \lar \emptyset$.}
\For{iteration $k = 1, 2, \cdots$}
\For{time step $t = 1, \cdots, T$}
\State{Observe the current state $s_t$.}
\State{Select the action $a_t$.}
\If{$\Vert a_t - \piE(s_t) \Vert^2 > \tau $ (or $a_t \ne \piE(s_t)$ for discrete action control) }
\State{Implement the action $\piE(s_t)$.}
\State{Receive the environment reward $r_t$ and observe the next state $s_{t+1}$.}
\State{Update $\gD^{o} \lar \gD^{o} \cup \{ (s_t, \piE(s_t), r_t, s_{t+1}) \}$.}
\Else
\State{Implement the action $a_t$.}
\State{Receive the environment reward $r_t$ and observe the next state $s_{t+1}$.}
\EndIf
\EndFor
\If{$\gD^{o}$ is not empty}
\If{Action space is continuous}
\State{Fine-tune the model by \eqref{eq:fine_tune_2} and \eqref{eq:fine_tune_1}.}
\Else
\State{Fine-tune the model by \eqref{eq:fine_tune_discrete}.}
\EndIf
\EndIf
\State{$\gD^{o} \lar \emptyset$.}
\EndFor
\end{algorithmic}
\end{algorithm}

\section{Experiments}
\label{sec:experiments}

In this section, we conduct experiments to verify the effectiveness of the proposed methods.

\subsection{Robotics Locomotion Control}

\begin{figure}[htbp]
    \centering
    \includegraphics[width=0.6\linewidth]{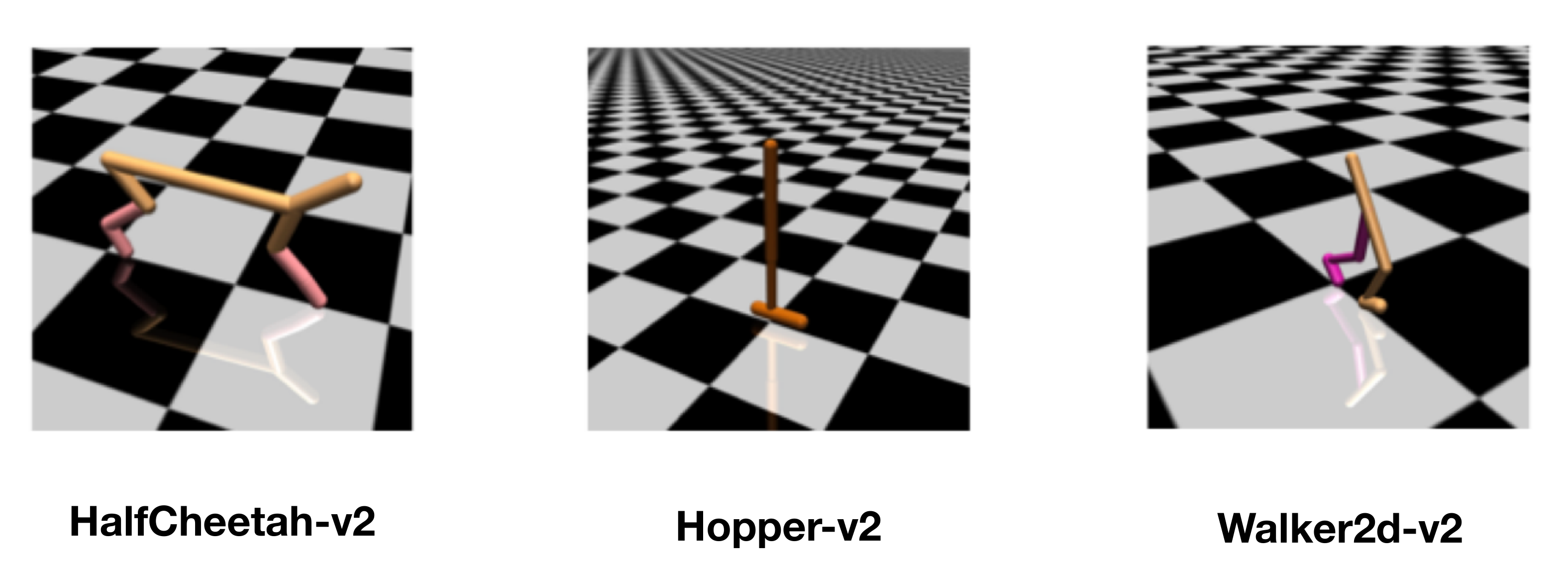}
    \caption{Illustration of robotics locomotion control, simulated by MuJoCo \cite{todorov2012mujoco}. }
    \label{fig:mujoco}
\end{figure}

This section focuses on three locomotion control tasks: HalfCheetah-v2, Hopper-v2, and Walker2d-v2, as described in \cite{duan2016benchmarking}. These tasks aim to train a robot to perform human-like locomotion behaviors, using joint angle and velocity information as states and low-level motor controls as actions. Refer to \cref{fig:mujoco} for a visual representation of the locomotion tasks.

For illustrative purpose, we employ an online RL algorithm, specifically the SAC algorithm \cite{haarnoja2018sac}, to obtain an expert policy. Note that SAC is run for 1 million steps. The replay buffer from SAC serves as our offline dataset.

Numerous approaches exist for training offline RL models, as seen in \cite{fujimoto2019offpolicy, kumar2020conservative, wu2020behavior, kostrikov2021offline}. For this study, we opt to use the CQL algorithm \citep{kumar2020conservative} due to its simplicity in implementation, although other methods can also be considered. CQL incorporates a penalty term on out-of-distribution actions during training, resulting in five offline RL models with different scales of this penalty term. The performance of these models is presented in \cref{tab:summary_mujoco} in the Appendix. In computing the online score, we appropriately scaled the environment reward and penalty terms for the considered tasks; refer to \eqref{eq:online_score}:
\begin{align*}
    \texttt{HalfCheetah-v2}: \quad &\alpha_1 = 1/8500, \alpha_2 = 1/1000, \\
    \texttt{Hopper-v2}: \quad &\alpha_1 = 1/3500, \alpha_2 = 1/1000, \\
    \texttt{Walker2d-v2}: \quad &\alpha_1 = 1/4000, \alpha_2 = 1/1000.
\end{align*}
The tolerance parameter $\tau = 0.09$ and the maximum trajectory length $T = 1000$.

\subsubsection{Online Model Selection}
\label{sec:experiment_online_model_selection}

Using the five trained offline models, we implement model selection during the online deployment phase with \cref{algo:ucb}. For \cref{algo:ucb}, we set the hyper-parameter $\beta$ to $1$. Our baselines include the following:

\begin{itemize}
\item Highest Q: This method selects the model with the highest Q-value function trained by offline datasets, which has been considered in prior work \citep{garg2020batch}.
\item Random Ensemble: This method randomly selects a model from the candidate models to deploy. This method is not adaptive during the online phase.
\end{itemize}
In addition, we also consider the oracle, which directly selects the model that can maximize the online score. Note that this method cannot be used in practice, and its performance serves as the upper limit for all methods.

\begin{figure*}[htbp]
    \centering
    \includegraphics[width=\linewidth]{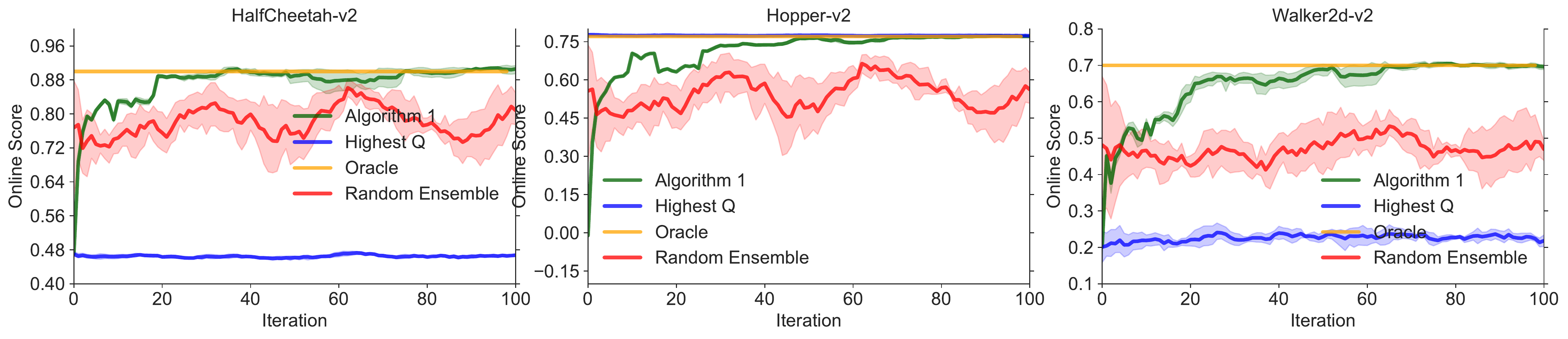}
    \caption{Online score (the higher the better) of \cref{algo:ucb} for the robotics locomotion control tasks. Solid lines correspond to the mean and shaded regions correspond to the 95\% confidence interval over $5$ random seeds (same as other figures). }
    \label{fig:ucb_score}
\end{figure*}

We present the numerical results in terms of online scores in \cref{fig:ucb_score}. Notably, we observe that with no more than 100 iterations, the performance of \cref{algo:ucb} is close to the optimal score. On the other hand, other methods like Highest Q and Random Ensemble do not perform well. The performance of the trained policies is reported in \cref{tab:online_selection_mujoco}, where the model selected by \cref{algo:ucb} achieves a high online score.

\begin{table*}[t]
\centering
\caption{Performance of online model selection algorithms for three locomotion control tasks. Note that the metric of human disagreement measures the number of actions on which the human expert disagrees with the model, as defined in \eqref{eq:online_score} and \eqref{eq:online_score_1}.  Here digits correspond to the averaged results, the symbol $\pm$ corresponds to the stand deviation of 5 experiments with different random seeds, and the symbol $\uparrow$ indicates that higher values are better, while the symbol $\downarrow$ indicates the opposite (same as other tables). }
\label{tab:online_selection_mujoco}
\begin{tabular}{@{}l|l|lll@{}}
\toprule
                             &                              & Environment Return ($\uparrow$) & Human Disagreement ($\downarrow$) & Online Score ($\uparrow$) \\ \midrule
\multirow{4}{*}{HalfCheetah-v2} & Highest Q  & \meanstd{8682}{34}               & \meanstd{554}{7}               & \meanstd{0.46}{0.01}        \\
                             & Random Ensemble     &     \meanstd{8246}{147}           & \meanstd{162}{67}               & \meanstd{0.78}{0.10}         \\
                             & Algorithm 1      & \meanstd{8308}{136}               & \meanstd{72}{11}                & {\meanstd{0.91}{0.00}}       \\
                             & Oracle     &  8292               & 59               & 0.90         \\
                            \midrule
\multirow{4}{*}{Hopper-v2}   & Highest Q     & \meanstd{3387}{1}                 & \meanstd{194}{1}               & {\meanstd{0.77}{0.00}}        \\
                             & Random  Ensemble   & \meanstd{2756}{465}               & \meanstd{225}{41}              & \meanstd{0.54}{0.17}         \\
                             & Algorithm 1     & \meanstd{3390}{6}               & \meanstd{198}{6}              & \textbf{\meanstd{0.77}{0.00}}         \\
                             & Oracle   & 3387               & 193              & 0.77         \\
                             \midrule
\multirow{4}{*}{Walker2d-v2} & Highest Q   & \meanstd{3627}{325}                & \meanstd{688}{56}              & \meanstd{0.23}{0.02}        \\
                             & Random Ensemble   & \meanstd{3685}{274}                & \meanstd{452}{96}              & \meanstd{0.40}{0.10}         \\
                             & Algorithm 1    & \meanstd{4100}{46}                & \meanstd{329}{9}              & {\meanstd{0.70}{0.00}}         \\
                             & Oracle  & 4106                & 328             & 0.70           \\
\toprule
\end{tabular}
\end{table*}

\subsubsection{Online Model Fine-Tuning}

\begin{table*}[t]
\centering
\caption{Performance of online fine-tuning algorithms for three locomotion control tasks.}
\label{tab:online_fine_tune_mujoco}
\begin{tabular}{@{}l|l|lll@{}}
\toprule
                             &                              & Environment Return ($\uparrow$) & Human Disagreement ($\downarrow$) & Online Score ($\uparrow$) \\ \midrule
\multirow{2}{*}{HalfCheetah-v2} & Without Fine-tuning  & \meanstd{7362}{46}               & \meanstd{93}{7}               & \meanstd{0.76}{0.00}        \\
                             & Algorithm 2      & \meanstd{7353}{90}               & \meanstd{67}{10}                & {\meanstd{0.80}{0.00}}       \\
                            \midrule
\multirow{2}{*}{Hopper-v2}   & Without Fine-tuning    & \meanstd{3368}{2}                 & \meanstd{264}{7}               & {\meanstd{0.70}{0.00}}        \\
                             & Algorithm 2   & \meanstd{3374}{2}               & \meanstd{81}{5}              & \meanstd{0.88}{0.00}         \\
                             \midrule
\multirow{2}{*}{Walker2d-v2} &  Without Fine-tuning   & \meanstd{3787}{39}                & \meanstd{253}{12}              & \meanstd{0.69}{0.00}        \\
                             & Algorithm 2    & \meanstd{3843}{25}                & \meanstd{118}{14}              & {\meanstd{0.84}{0.00}}         \\  
                             \toprule
\end{tabular}
\end{table*}

In this part, we consider the online model fine-tuning approaches for deploying RL models as introduced in \cref{sec:online_model_fine_tuning}. Without loss of generality, we select the offline model with the worst performance to fine tune (refer to \cref{tab:summary_mujoco} in Appendix for the detailed performance of offline models).

\begin{figure*}[htbp]
    \centering
    \includegraphics[width=\linewidth]{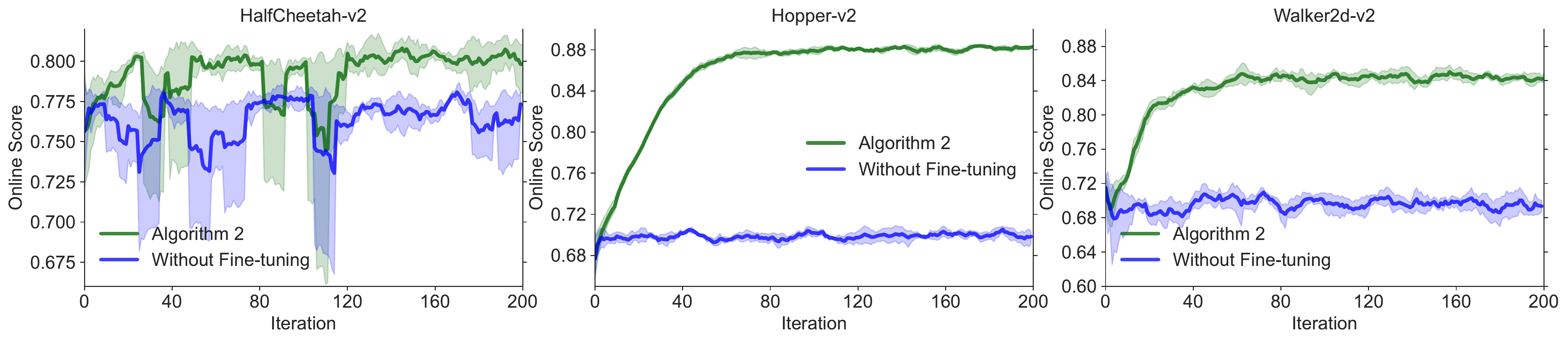}
    \caption{Online score of \cref{algo:fine_tune} for the robotics locomotion control tasks. }
    \label{fig:fine_tune_mse}
\end{figure*}

We report the numerical results in \cref{fig:fine_tune_mse} and \cref{tab:online_selection_mujoco}. From \cref{fig:fine_tune_mse}, we see that by minimizing the mean squared error to fine-tune the RL model further, its online performance can be improved, compared with that without fine-tuning. This validates the effectiveness of \cref{algo:fine_tune}.

\subsection{Traffic Light Control}

\begin{figure}[htbp]
    \centering
    \includegraphics[width=0.3\linewidth]{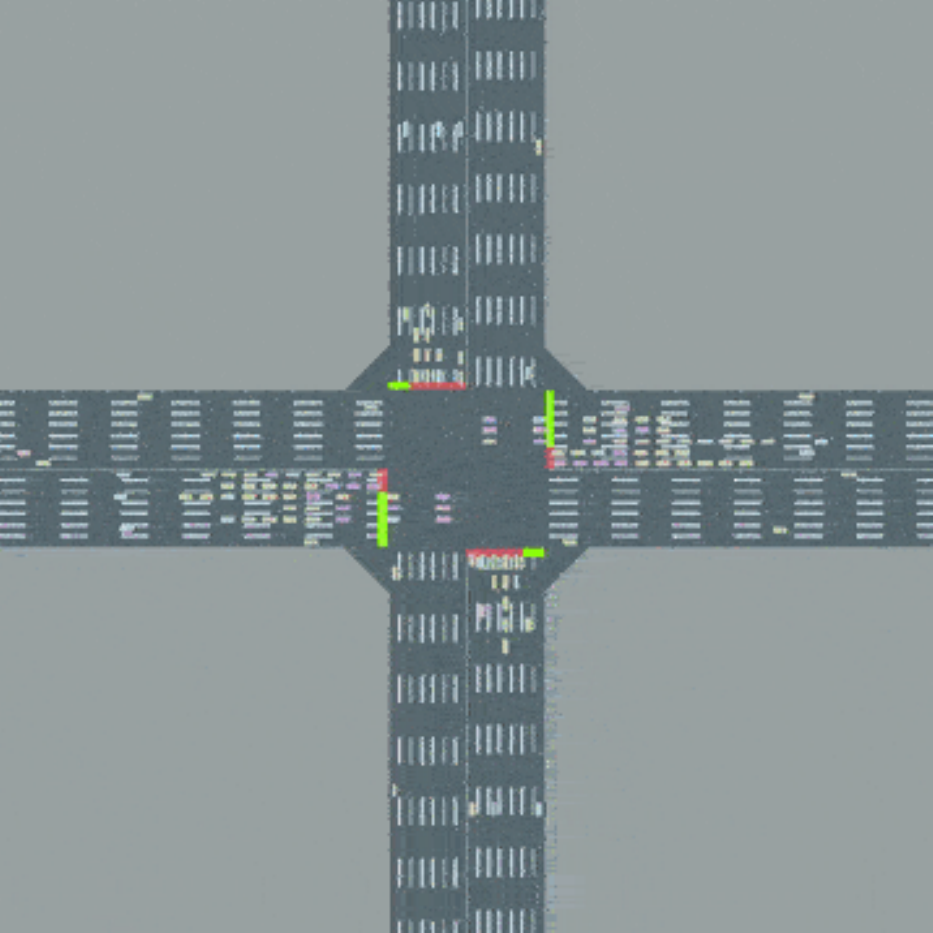}
    \caption{Illustration of the traffic light control, simulated by CityFlow \cite{zhang2019cityflow}.}
    \label{fig:traffic_light}
\end{figure}

This section considers a traffic light control task, as shown in \cref{fig:traffic_light}. The traffic light at the roadway is the agent that needs to be controlled, while traffic flows serve as the environment. The states of the system include the queue length (i.e., the number of vehicles in incoming lanes) and the current phase (i.e., the movement signal of the traffic light, such as a green light on the west-east). The goal is to minimize queue length and avoid congestion by adaptively selecting the movement signal. We use real traffic data from Hangzhou, China, which was provided by the TSCC competition\footnote{\url{https://github.com/tianrang-intelligence/TSCC2019}}.

Since the action space for traffic light control is discrete and finite, we consider using DQN-based approaches \cite{mnih2015human}. To obtain an expert policy, we first train a double DQN agent \cite{hasselt2016double} for 100 iterations. We then collect an offline dataset of 500K samples by using $\epsilon$-greedy ($\epsilon=0.2$) to roll out the expert policy. We use CQL \cite{kumar2020conservative} to train offline RL models with different penalty scales, resulting in five models. Their performance is summarized in \cref{tab:summary_traffic} in Appendix. We set $\alpha_1 = 1/1000$ and $\alpha_2 = 1/1800$ for this traffic light control task.

\subsubsection{Online Model Selection}

\begin{figure}[htbp]
    \centering
    \includegraphics[width=0.45\linewidth]{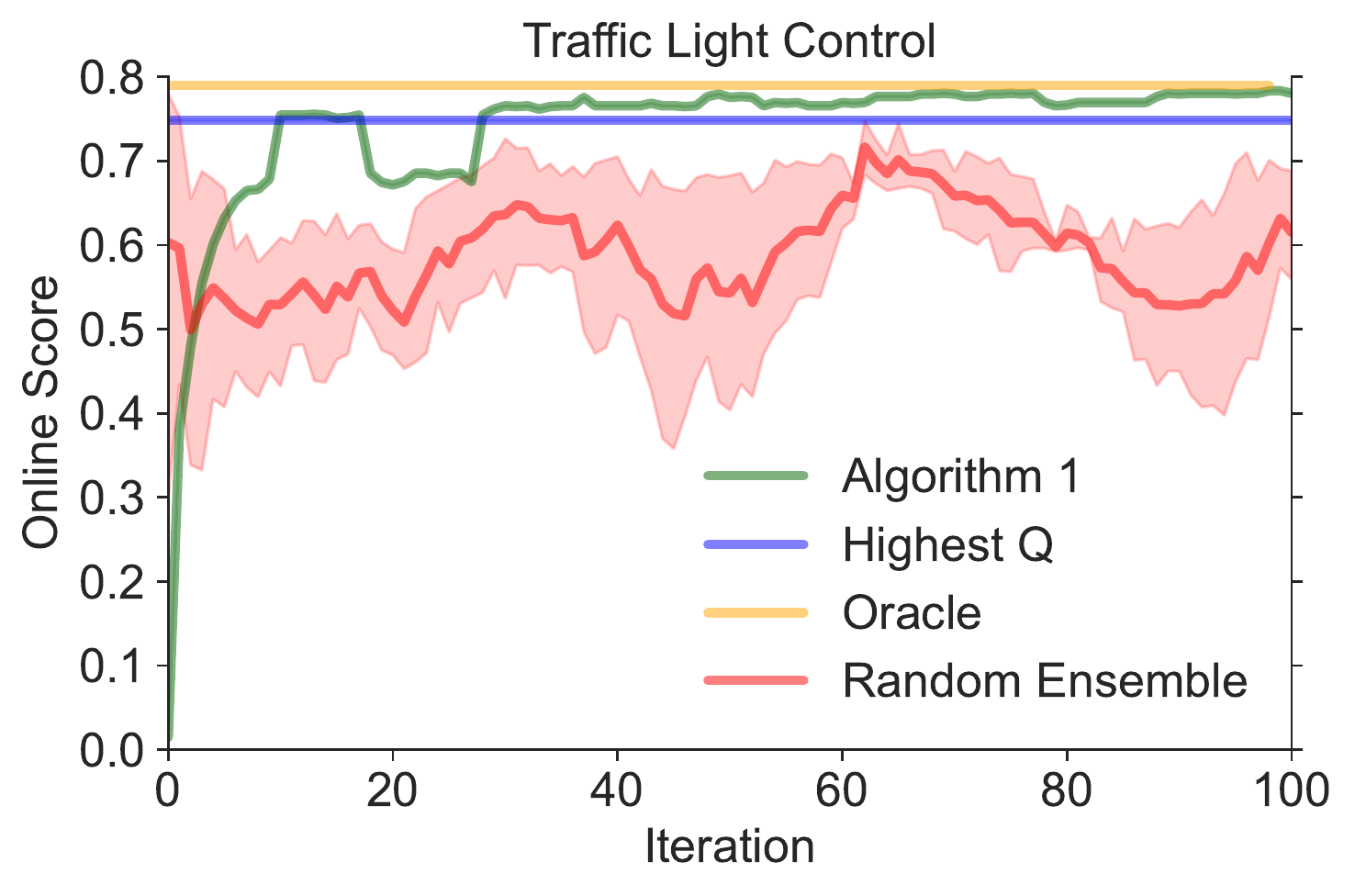}
    \caption{Online score of \cref{algo:ucb} for the traffic light control task. }
    \label{fig:traffic_light_ucb_score}
\end{figure}

In this section, we use \cref{algo:ucb} to select offline RL models during the online deployment phase, with a chosen hyper-parameter $\beta$ of $1$. We consider the same baselines as in \cref{sec:experiment_online_model_selection} and present the online score curve in \cref{fig:traffic_light_ucb_score}. Our observation is consistent with the previous section: after 100 iterations, the performance of \cref{algo:ucb} is comparable to the optimal one. We report the detailed performance of trained policies in \cref{tab:online_selection_traffic}.

\begin{table}[t]
\centering
\caption{Performance of online model selection algorithms for the traffic light control task.  }
\label{tab:online_selection_traffic}
\begin{tabular}{c|lll@{}}
\toprule
                 & Environment Return ($\uparrow$) & Human Disagreement ($\downarrow$) & Online Score ($\uparrow$) \\ \midrule
Highest Q        & \meanstd{908}{0}               & \meanstd{320}{0}              & \meanstd{0.75}{0.00}          \\ 
Random Ensemble  & \meanstd{898}{11}               & \meanstd{565}{199}               & \meanstd{0.57}{0.03}         \\  
Algorithm 1      & \meanstd{916}{3}               & \meanstd{270}{17}               & \meanstd{0.78}{0.00}          \\  
Oracle    & 913               & 270               & 0.79         \\ \toprule
\end{tabular}
\end{table}

\subsubsection{Online Model Fine-Tuning}

\begin{figure}[htbp]
    \centering
    \includegraphics[width=0.45\linewidth]{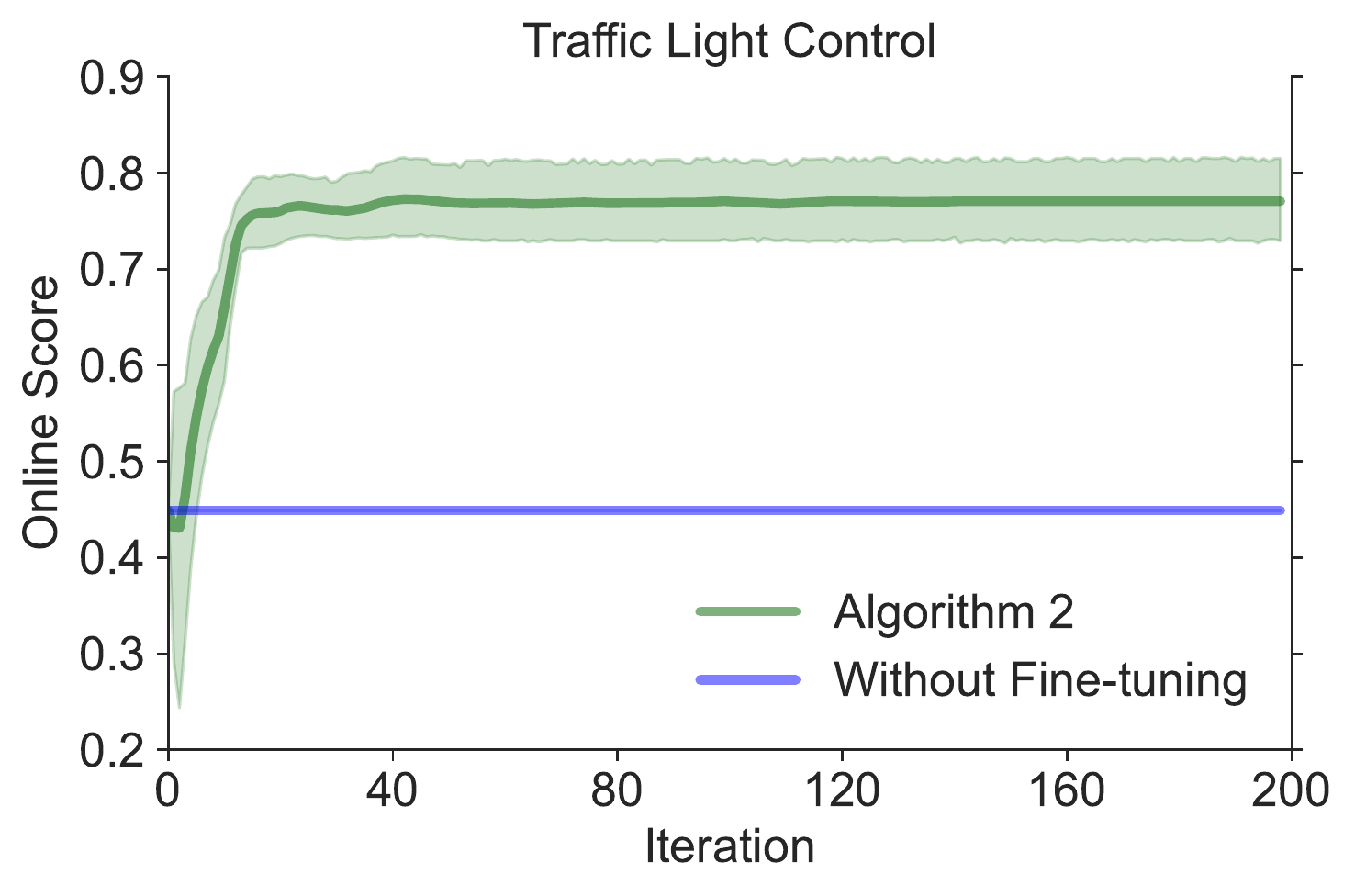}
    \caption{Online score of \cref{algo:fine_tune} for the traffic light control task. }
    \label{fig:traffic_light_fine_tune_score}
\end{figure}

In this part, we try to fine-tune the trained RL model in the online deployment phase. Again, we select the model withe the worst performance to fine-tune. The hyperparameter $\Delta = 1$ in \eqref{eq:fine_tune_discrete} is used in experiments.  We visualize the online score curve in \cref{fig:traffic_light_fine_tune_score}. We observe that without fine-tuning, the performance of the deployed RL model is poor. However, by fine-tuning this model via \cref{algo:fine_tune} with $100$ iterations, its performance can be significantly improved. The detailed performance of trained policies is reported in \cref{tab:online_fine_tune_traffic}.

\begin{table}[t]
\centering
\caption{Performance of online fine-tuning algorithms for the traffic light control task.}
\label{tab:online_fine_tune_traffic}
\begin{tabular}{c|lll@{}}
\toprule
                 & Environment Return ($\uparrow$) & Human Disagreement ($\downarrow$) & Online Score ($\uparrow$) \\ \midrule
Without Fine-tuning       & \meanstd{906}{0}               & \meanstd{457}{0}              & \meanstd{0.45}{0.00}          \\  
Algorithm 2  & \meanstd{928}{0}               & \meanstd{147}{0}               & \meanstd{0.77}{0.00}         \\ \toprule
\end{tabular}
\end{table}

\section{Conclusion}
\label{sec:conclusion}

This manuscript explores effective deployment strategies for offline reinforcement learning (RL) models in the online phase, leveraging human feedback. Two approaches are proposed: model selection and fine-tuning. Experimental results demonstrate the effectiveness of these methods in achieving high online performance.

It should be noted that this work only considers scenarios where the human expert policy and environment are static. In some applications, these factors may change over time, and more sophisticated deployment methods may be required in the future.

\bibliographystyle{abbrvnat}
\bibliography{reference}  

\appendix

\section{Experiment Details}

Our algorithm implementation of SAC and CQL is based on the \texttt{tianshou} framework\footnote{\url{https://github.com/thu-ml/tianshou}}. We use $\lambda$ to indicate the scale of the penalty term in CQL. The performance of offline models is reported in \cref{tab:summary_mujoco} for robotics locomotion control tasks and in \cref{tab:summary_traffic} for the traffic light control task. We note that results in \cref{tab:online_selection_mujoco}, \cref{tab:online_fine_tune_mujoco}, \cref{tab:online_selection_traffic}, and \cref{tab:online_fine_tune_traffic} are based on the average of last 10 iterations.

\begin{table*}[t]
\centering
\caption{Performance of trained offline models by CQL for robotics locomotion control tasks.  Statistics are estimated from 20 evaluation episodes (same with \cref{tab:summary_traffic}). Note that these statistics cannot be obtained in the offline training phase.}
\label{tab:summary_mujoco}
\begin{tabular}{@{}l|l|ccc@{}}
\toprule
                             &                              & Environment return ($\uparrow$) & Human Disagreement ($\downarrow$) & Online Score ($\uparrow$) \\ \midrule
\multirow{4}{*}{HalfCheetah-v2} & Model 1 ($\lambda = 0$)   & 8731               & 559               & 0.47        \\
                             & Model 2 ($\lambda = 1$)      & 8732               & 131               & 0.90         \\
                             & Model 3 ($\lambda = 5$)      & 8292               & 59                & 0.92         \\
                             & Model 4 ($\lambda = 10$)     & 8076               & 49                & 0.90         \\
                             & Model 5 ($\lambda = 100$)    & 7328              & 104              & 0.76         \\
                            \midrule
\multirow{4}{*}{Hopper-v2}   & Model 1 ($\lambda = 0$)     & 39                 & 22               & 0.00         \\
                             & Model 2 ($\lambda = 1$)     & 3450               & 266              & 0.72         \\
                             & Model 3 ($\lambda = 5$)     & 3387               & 193              & 0.77         \\
                             & Model 4 ($\lambda = 10$)    & 3309               & 299              & 0.65         \\
                             & Model 5 ($\lambda = 100$)   & 3312               & 303              & 0.64         \\
                             \midrule
\multirow{4}{*}{Walker2d-v2} & Model 1 ($\lambda = 0$)    & 3640                & 699              & 0.21         \\
                             & Model 2 ($\lambda = 1$)    & 4106                & 328              & 0.70         \\
                             & Model 3 ($\lambda = 5$)    & 3726                & 695              & 0.24         \\
                             & Model 4 ($\lambda = 10$)   & 3767                & 302              & 0.64         \\
                             & Model 5 ($\lambda = 100$)  & 3667                & 360              & 0.56         \\
\toprule
\end{tabular}
\end{table*}

\begin{table}[t]
\centering
\caption{Performance of trained offline models by CQL for the traffic light control task.}
\label{tab:summary_traffic}
\begin{tabular}{l|ccc@{}}
\toprule
                                                      & Environment return ($\uparrow$) & Human Disagreement ($\downarrow$) & Online Score ($\uparrow$) \\ \midrule
Model 1  ($\lambda = 0$)     & 835               & 1641              & 0.01          \\ \hline 
Model 2  ($\lambda = 1$)     & 908               & 320               & 0.75         \\  \hline
Model 3  ($\lambda = 5$)     & 906               & 457               & 0.68          \\  \hline
Model 4  ($\lambda = 10$)    & 913               & 270               & 0.78          \\ \hline
Model 5  ($\lambda = 100$)   & 918               & 262               & 0.79         \\ \toprule
\end{tabular}
\end{table}

\end{document}